\documentclass[letterpaper, 10 pt, conference]{ieeeconf}  

\IEEEoverridecommandlockouts                              

\usepackage{booktabs}
\usepackage{graphicx}                                                         
\usepackage{float} 
\usepackage[table]{xcolor} 
\usepackage{multirow}
\usepackage{multicol}
\usepackage{amsmath}  

\usepackage{stfloats}  

\usepackage {hyperref}  

\title{\LARGE \bf
FISRF: Flexible Instructional Scaffolding Reflection Framework for Robotics Task Planning with Large Language Model
}

\author{QingMiao Liang$^{1\dag}$, Jiatao Zhang$^{2\dag}$, Runxin Zou$^{2}$, Yufan Song$^{2}$, Wei Song$^{1,2\star}$, YuHua Zheng$^{1\star}$
\thanks{$^{1}$Hangzhou Institute for Advanced Study, University of Chinese Academy of Sciences, Hangzhou, China}
\thanks{$^{2}$Zhejiang University, Hangzhou, China}%
\thanks{$^{\star}$Corresponding authors' emails: 
\texttt{weisong-rob@zju.edu.cn} and 
\texttt{zhengyuhua@ucas.ac.cn}.}
\thanks{$^{\dag}$ Contribute equally to this work.}
}

\begin{document}

\maketitle
\thispagestyle{empty}
\pagestyle{empty}

\begin{abstract}

Large Language Models (LLMs) are widely adopted for reflective error correction in task planning; however, existing approaches primarily rely on autonomous exploration, rendering them prone to accumulating erroneous experiences during early interactions with new environments. To address this limitation, we propose the Flexible Instructional Scaffolding Reflection Framework (FISRF), an agent--mentor architecture that estimates the robot's Zone of Proximal Development (ZPD) in task planning and adaptively selects reflection strategies and mentor guidance, thereby facilitating the accumulation of effective error-correction experience. Experiments in ALFWorld and on real-world robotic platforms demonstrate that FISRF reduces the average number of error-correction rounds by 37\% and redundant reasoning by 25.1\%, while improving the overall task success rate by 8\%. Project webpage : https://fisrfplanning.github.io/


\end{abstract}

\section{INTRODUCTION}
\begin{figure}[t!]
    \centering
    \includegraphics[width=0.50\textwidth]{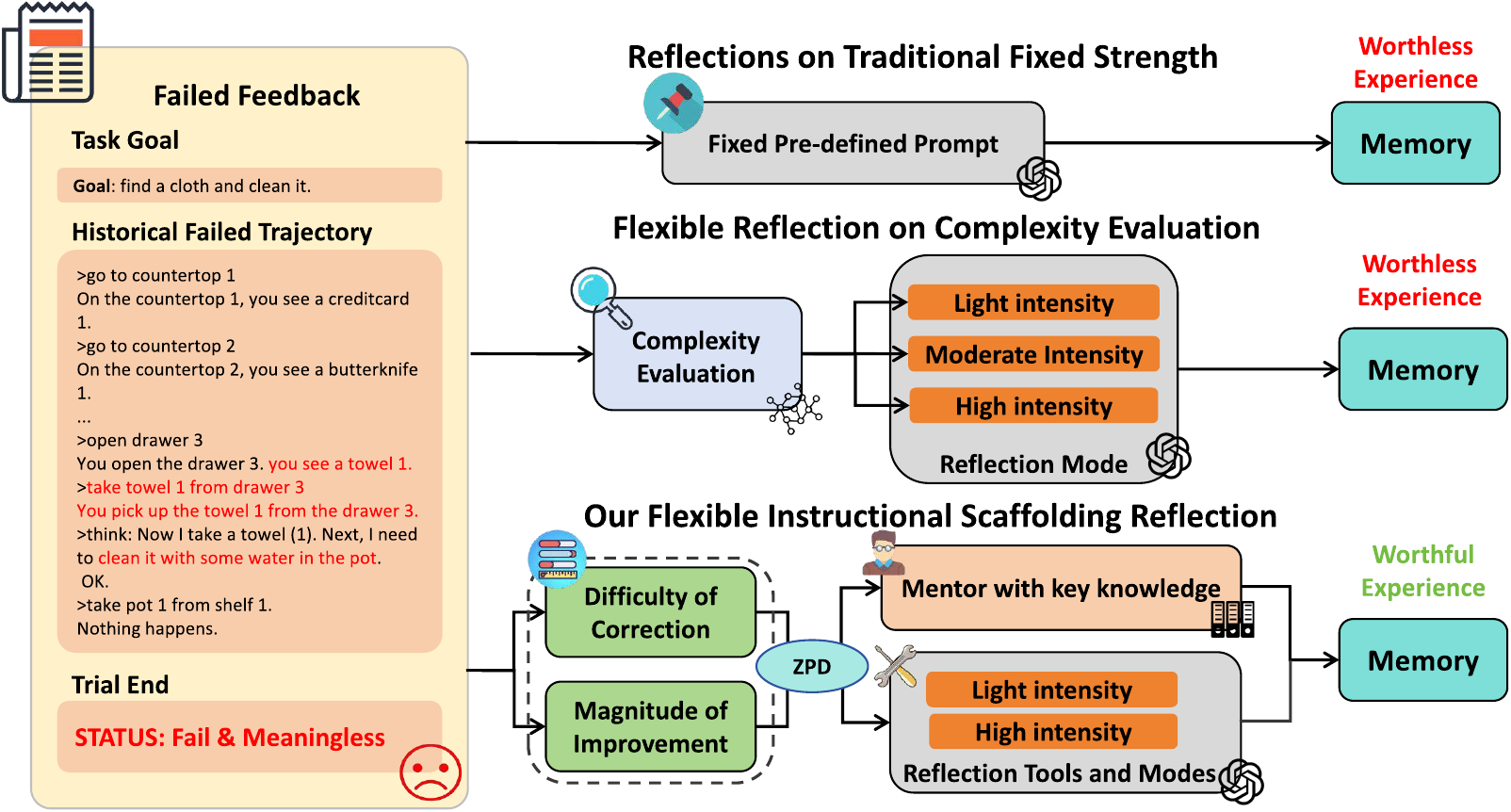}
    \caption{Compared with existing approaches, FISRF adaptively determines whether to solicit mentor guidance and selects appropriate reflection strategies according to the difficulty of correction and the magnitude of improvement, in each trial, thereby effectively preserving and leveraging valuable experience for subsequent correction.}
    \label{FISRF_fig1}
 \vspace{-5mm}
\end{figure}

Recent advances in robot task planning increasingly leverage large language models (LLMs) to generate coherent multi-step plans~\cite{singh2023progprompt, zhang2024fltrnn}. Despite strong performance on complex tasks, LLM-based planners remain susceptible to hallucinations and logical inconsistencies~\cite{huang2024understanding}. To alleviate these limitations, reflection 
. RETROFORMER~\cite{yao2023retroformer} and Reflexion~\cite{shinn2024reflexion} extend the ReAct~\cite{yao2022react} framework by incorporating reflective feedback through textual gradients. Another research line~\cite{yao2024tree,zhang2024agent,zhou2023language} integrates tree-search strategies with reflection loops to support more deliberate exploration. More recent systems, such as Expel~\cite{zhao2024expel} and AutoManual~\cite{chen2024automanual}, construct environment-specific rule libraries that enable structured retrospective error analysis. Furthermore, FCRF~\cite{song2025fcrf} and Thinkless~\cite{fang2025thinkless} quantify task complexity to adaptively adjust reflection intensity (see Figure~\ref{FISRF_fig1}), thereby transforming reflection from a fixed procedure into a flexible mechanism.

Most existing reflection methods rely on robots operating autonomously under predefined reflection procedures, enabling effective experience accumulation in familiar or structured tasks. However, when robots enter new environments, they often lack effective experience to handle unfamiliar problems and may be unable to correct seemingly simple errors. These early-stage errors can accumulate, leading to numerous ineffective attempts and resulting in reduced efficiency and degraded performance.(Figure~\ref{FISRF_fig1}) We therefore ask: \textit{How can robots rapidly accumulate effective experience in new environments while maintaining a certain degree of autonomy?}

To address these limitations, we build on the theory of \textbf{Instructional Scaffolding} from educational psychology~\cite{VanDerStuyf2002ScaffoldingTeaching}. This theory suggests that a \textbf{learner’s problem-solving ability improves most effectively when guidance is provided within their Zone of Proximal Development (ZPD)}—the gap between what can be accomplished independently and what can be achieved with appropriate guidance. When persistent obstacles arise during independent exploration, targeted instruction that addresses core difficulties facilitates knowledge internalization and transfer, enabling more robust mastery of complex skills.

This raises a key challenge:  
\textbf{the lack of a quantitative and operational formulation of the robot’s Zone of Proximal Development in task planning.} Addressing this problem requires formalizing and automating the dynamic guidance mechanisms traditionally provided by human teachers. Specifically, the robot’s task performance must be quantitatively and adaptively modeled to establish a capability baseline and delineate its developmental boundaries. Based on this baseline, appropriate guidance strategies should be automatically generated. Moreover, the framework must progressively withdraw external support and consolidate assisted experience into autonomous, generalizable competence, thereby forming a closed loop from guided learning to independent execution.

Motivated by the aforementioned challenge, we propose the \textbf{Flexible Scaffolding Reflection Framework} (FISRF), an online approach that provides adaptive guidance to large language models based on task performance. FISRF is grounded in instructional scaffolding theory and adopts an agent–mentor architecture. The \textbf{Agent module} consists of three components: 1) an \textbf{Actor} LLM that generates action sequences; 2) an \textbf{Evaluator} LLM that determines ZPD entry by dynamically modeling \textbf{two ZPD factors}: difficulty of correction ($DC$) and magnitude of improvement ($MI$); and 3) \textbf{Self-Reflection} LLMs that perform experience summarization and lesson extraction.  
The \textbf{Mentor module} is implemented as a retrieval-augmented generation (RAG) component with a structured knowledge base. It retrieves task-relevant context and generates state-conditioned guidance grounded in the agent’s reflective feedback. After task execution, the agent evaluates its performance to determine ZPD entry. If the condition is satisfied, it requests mentor guidance and integrates the retrieved guidance with self-reflection for subsequent planning, thereby reducing redundant trials and enabling efficient, adaptive experience accumulation.

To assess the effectiveness of the proposed framework, we conduct experiments on the ALFWorld~\cite{shridhar2020alfworld} household planning benchmark. The tasks in the dataset are systematically organized into six categories based on their operational characteristics. Empirical results indicate that our approach attains superior performance across all task types.
In summary, our contributions are fourfold:
\begin{enumerate}
\item We propose FISRF, an adaptive agent–mentor reflection framework that balances autonomous exploration and external guidance.
\item We operationalize the ZPD for robotic task planning using difficulty of correction and magnitude of improvement, enabling dynamic selection of reflection strategies.
\item We validate FISRF in ALFWorld and real-world robotic tasks, demonstrating improved task success, correction efficiency, and reflection flexibility.
\end{enumerate}

\section{RELATED WORKS}

\subsection{LLMs for Task Planning}

Recent studies have leveraged the reasoning and long-horizon decision-making capabilities of LLMs for task planning. LID~\cite{li2022pre} formulates observations, goals, and interaction histories as sequential representations for policy prediction, while ReAct~\cite{yao2022react} interleaves reasoning and action generation. Code-based methods, including CodeAsPolicy~\cite{liang2023code}, ProgPrompt~\cite{singh2023progprompt}, and AdaPlanner~\cite{sun2024adaplanner}, further represent plans as executable programs to improve structural consistency and interpretability. In embodied robotics, SayCan~\cite{ahn2022can} grounds language-based reasoning in environmental affordances, whereas LLMPlanner~\cite{song2023llm,wake2023chatgpt} extends LLM-based planning to embodied decision-making tasks.

\subsection{LLM Agents for Self-Reflection}

Building upon early one-shot planning strategies, subsequent research introduces self-reflection mechanisms that incorporate feedback for multi-step planning and iterative error correction in LLM-based agents. For instance, extending the ReAct framework, Reflexion~\cite{shinn2024reflexion} enables agents to analyze prior failures using environmental feedback and refine their plans in subsequent trials. Expel~\cite{zhao2024expel} further develops this paradigm by introducing an offline learning stage, where agents accumulate experience across tasks through trial-and-error interactions.
AutoManual~\cite{chen2024automanual} constructs a structured representation of the environment to facilitate effective multi-task planning. Additionally, several approaches~\cite{yao2024tree,zhang2024agent,zhou2023language} integrate tree search with reflective mechanisms to improve solution quality during task execution. Building on the fixed-intensity reflection paradigm described above, FCRF~\cite{song2025fcrf} introduces an adaptive reflection strategy conditioned on task complexity, thereby improving reflection effectiveness and overall task-planning performance.

\section{PRELIMINARIES}

\subsection{Planning Framework}

A task can be formally represented as a tuple
$\langle {G}, {S}, {O}, {A}, {T} \rangle$,
where ${G}$ specifies the goal condition, ${S}$ denotes the state space,
${O}$ corresponds to the observation space, ${A}$ is the action set,
and ${T}: {S} \times {A} \rightarrow {S}$ defines the state transition dynamics induced by action execution.

The objective of planning is to identify a policy—typically instantiated as an action sequence—that guides the system from an initial state to a target state satisfying ${G}$.
In LLM-based planning frameworks, such structured task elements are commonly encapsulated in a natural language description $D$ provided to the model.

\subsection{Self-Reflection Process in LLM-based Planning}
Following Reflexion~\cite{shinn2024reflexion}, we formalize Reflection as an iterative optimization procedure guided by environmental feedback. At each trial $t$, the planner interacts with the environment to generate a trajectory $\tau_t$, and obtains a scalar evaluation $r_t = M_{E}(\tau_t)$. The reflection module maps the pair $(\tau_t, r_t)$ into a linguistic feedback signal $sr_t = M_{\text{ref}}(\tau_t, r_t),$
which is stored in memory $\texttt{mem}$ and leveraged to update the policy in subsequent trials. The iterative loop $(\tau_t, r_t, sr_t)$ continues until the environment feedback indicates that the task has been successfully completed.

\begin{figure*}[!htp]
    \centering
    \includegraphics[width=0.7\textwidth]{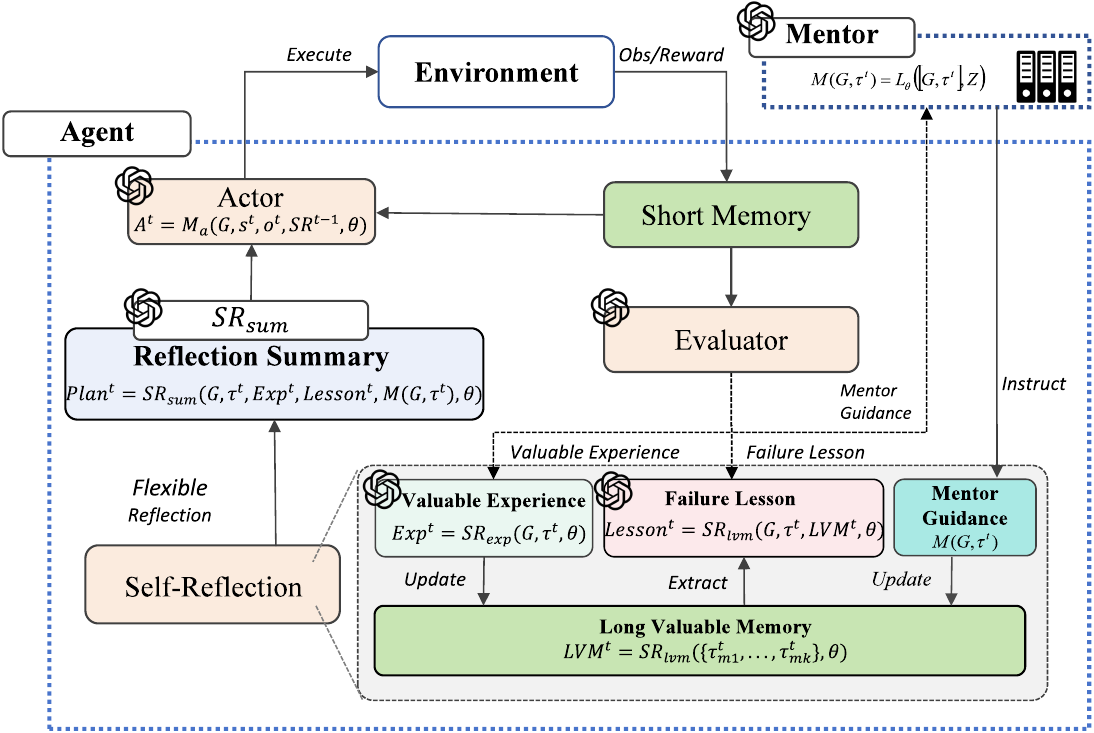}
    \caption{The framework of FISRF. The planning process is executed by the Actor, the evaluation process is executed by Evaluator, the reflection process is executed by the Self-Reflection, the difficulty of correction and magnitude of improvement are assessed, according to the assessment results, determine whether to seek mentor guidance and implement appropriate reflection strategies. By integrating insights from autonomous exploration with mentor guidance, our framework inform subsequent plan. The resulting reflections and planning trajectories are stored in a memory module for long-term retrieval.}
    \label{framework}
 \vspace{-5mm}
\end{figure*}

\section{METHODOLOGY}

In this section, we describe the Flexible Instructional Scaffolding Reflection Framework (FISRF). As shown in Figure, FISRF adopts an actor–mentor architecture that consists of four parts: an LLM as planning Actor, the Flexible Self-Reflection Process, the short-long memory management and the overall adaptive Evaluator model.We provide a detailed description of each of these models and subsequently elucidate their collaborative functioning within the FISRF framework.

\subsection{LLM as planning Actor}
Adopting the architectural paradigm of the Reflexion framework, the planning procedure is executed by an LLM-based $Actor$, denoted as $M_{a}$. At each iteration, $M_{a}$ processes the task goal, current environmental observations, and reflection summaries generated from previous failed trials. The model is prompted to generate a textual action sequence as output. The planning objective of $M_{a}$ is formulated as:

\begin{equation}
A^{t}=M_{a}(G,s^{t},o^{t},SR^{t-1},\theta),
\label{Actor_LLM} 
\end{equation}
where $A^{t}=\{a_{1}^{t},\dots,a_{i}^{t}\}$ denotes the current action sequence, $G$ represents the task goal, $o^{t}$ represents the environmental observations at trial $t$, $SR^{t-1}=\{sr^{t-k},\dots,sr^{t-1}\}$ denotes the set of self-reflection summaries from the previous $k$ trials under consideration, and $\theta$ represents the parameters of $M_{a}$.The action sequence generated by the Actor LLM $M_{a}$ is stored in the memory module $mem$ to support the subsequent self-reflection process, and this iterative procedure continues until the current task is successfully completed.

\subsection{The Short-Long Memory Management}
During task planning, maintaining contextual memory is essential for the LLM. However, incorporating all historical memory into the prompt introduces redundancy, increasing cognitive burden and potentially degrading performance.
Inspired by the long short-term memory mechanism in the Reflexion framework, we design a short-term memory module $sm$ and a long-term valuable memory module $lvm$ for structured memory management. The module $sm$ serves as an external buffer, comprising a trajectory management component ${T}^{t}$ and a reflection management component ${R}^{t}$, enabling systematic organization of task trajectories and reflection content.
Specifically, ${T}^{t}$ stores the task trajectory $\tau^{t}$ of the current trial $t$. In contrast, ${R}^{t}$ preserves all failure reflection records up to trial $t$, denoted as $\{r^{1},\dots,r^{t}\}$, for selective retrieval by the self-reflection module in subsequent trials.
The module $lvm$ stores key knowledge provided by the mentor, as well as valuable experiences accumulated through autonomous exploration, serving as a universal knowledge base for the task environment. This hierarchical memory mechanism mitigates the cognitive burden on the LLM while enhancing memory organization efficiency and reflection effectiveness.

\subsection{The RAG-Based Mentor Module}

The mentor module, denoted as $M$, employs a Retrieval-Augmented Generation (RAG) architecture that integrates a parameterized language model $L_\theta$ with a scenario-specific knowledge base ${D}$. The mentor knowledge base ${D}$ is constructed offline by extracting abstract task-level knowledge from successful examples, with instance-specific details removed. It is kept fixed during evaluation, and no validation or test examples are used. This hybrid approach enables the retrieval of task-relevant context to mitigate the limitations of static parametric knowledge. Given a task goal $G$ and the current trajectory $\tau^t$, the knowledge extraction and generation process is formulated as:
\begin{equation}
M(G, \tau^t) = L_\theta \left( [G, \tau^t], {Z} \right)
\label{eq:mentor_module}
\end{equation}
where ${Z} = \text{Top-}k_{z \in {D}} \left( \text{sim}(E_\phi(G, \tau^t), E_\phi(z)) \right)$ represents the set of the $k$ most relevant documents retrieved from ${D}$. Here, $E_\phi$ denotes the embedding model used to compute cosine similarity between the query and the knowledge base entries. The retrieved evidence ${Z}$ is concatenated with the query $[G, \tau^t]$ and fed into $L_\theta$ to produce grounded guidance.

\begin{figure*}[!htp]
    \centering
    \includegraphics[width=0.99\textwidth]{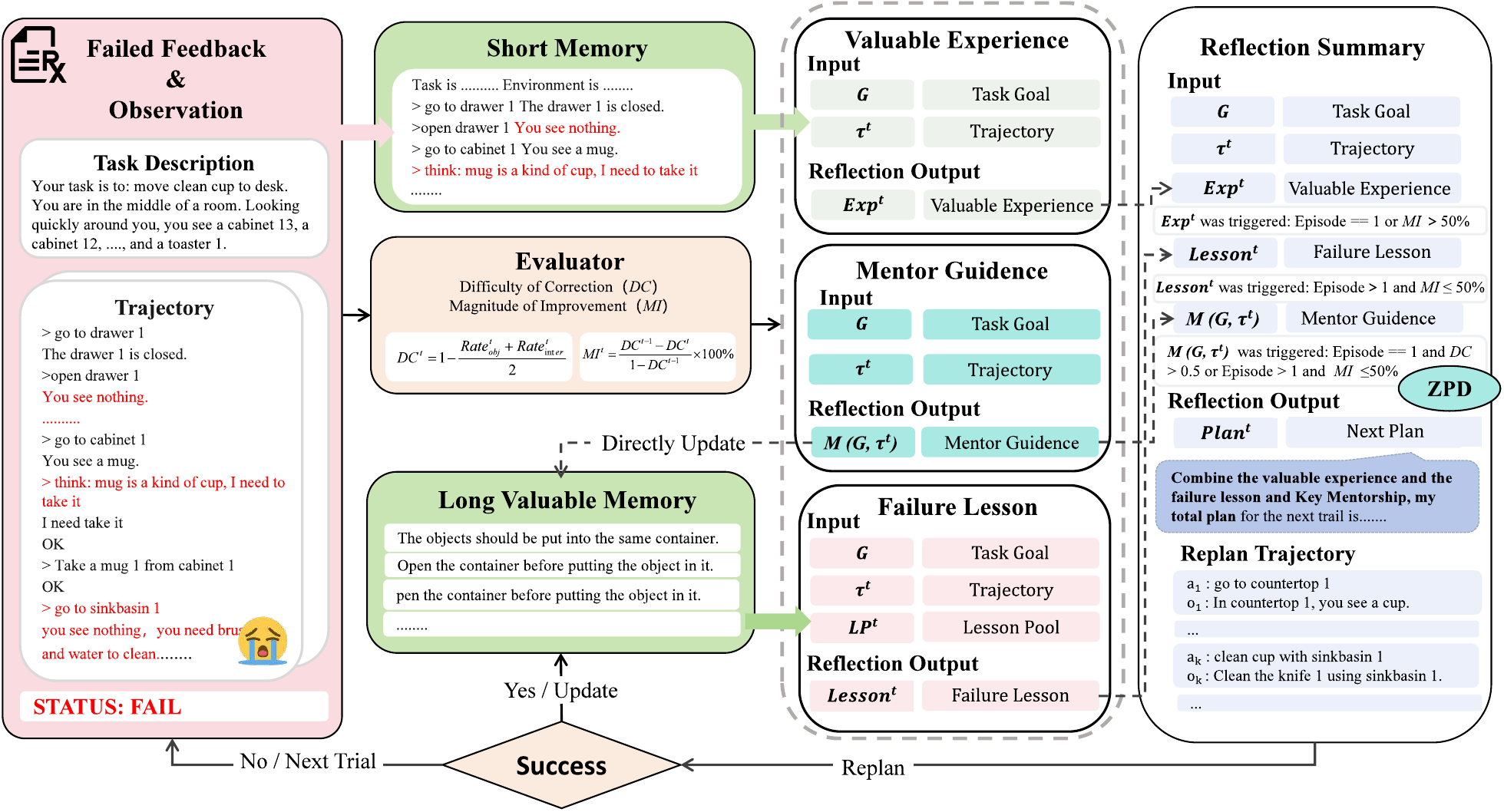}
    \caption{The detailed illustration of FISRF. The difficulty of correction and magnitude of improvement are first assessed, determining reflection strategies and whether to seek mentor guidance. After that, the detailed self-reflection process consists of the valuable experience summary process, the failure lesson summary process, Seek
     mentor guidance and the Reflection Summary process. Final output of the self-reflection process is a improved new plan for the next attempt.}
    \label{FISRF_fig3}
    \vspace{-5mm}
\end{figure*}

\subsection{The Self-Reflection Model}  
The Self-Reflection framework, denoted as $SR$, comprises three sub-modules: $SR_{exp}$, $SR_{lvm}$, and $SR_{sum}$. The reflection process is executed in three stages: (i) distilling high-value experiences from short-term memory, (ii) extracting failure-based lessons from long-term memory, and (iii) synthesizing a comprehensive reflection summary integrated with mentor-provided knowledge. The resulting output is stored in the short-term memory module $sm$, enabling the actor model $M_{a}$ to retrieve and leverage these insights for informed planning in subsequent trials.

\paragraph{The valuable experience summary model} This process is implemented in the submodule $SR_{exp}$. Specifically, $SR_{exp}$ analyzes the trial trajectory generated by $M_{a}$, identifies effective interaction attempts from unsuccessful trials, and distills them into reusable experience. The extracted insights are stored in memory to support subsequent reflection. The procedure can be summarized as follows:

\begin{equation}
Exp^{t}=SR_{exp}(G,\tau^{t},\theta),
\label{experience} 
\end{equation}
where $G$ represents the task goal, $\tau^{t}$ denotes the trajectory of the current trial $t$, and $\theta$ represents the parameters of $FSR_{exp}$, which are primarily instantiated through prompt design.

\paragraph{The failure lesson summary model} This model is completed by the submodule $SR_{lvm}$ and consists of two components: the preliminary maintenance of the long valuable memory and the extraction of failure  lessons.

During long valuable memory maintenance, $SR_{lvm}$ accesses existing cross-task trajectories online, summarizing universal failure lessons from all trajectories successfully corrected prior to the current episode, then incrementally updates ${lvm}$ with new lessons, formulated as:
\begin{equation}
LVM^{t}=SR_{lvm}(\{\tau^{t}_{m1},\dots,\tau^{t}_{mk}\},\theta),
\label{lesson_summarize} 
\end{equation}
where $LVM^{t}$ represents the up-to-date long valuable memory set up to trail $t$, $\{\tau^{t}_{m1},\dots,\tau^{t}_{mk}\}$ represents all $k$ successfully modified task trajectories in the current trail $t$, $\theta$ represents the parameters of $SR_{lvm}$, mainly embodies through prompt. 

During failure lesson extraction, $SR_{lvm}$ analyzes the current task trajectory against the task goal to retrieve the most relevant failure lesson from ${M}_l$, serving as external context for reflection. This lesson, combined with valuable experience $Exp^{t)}$, informs subsequent reflection content generation, formulated as:
\begin{equation}
Lesson^{t}=SR_{lvm}(G,\tau^{t},LVM^{t},\theta),
\label{lesson_extract} 
\end{equation}
where $G$ represents the task goal, $\tau^{t}$ represents the trajectory of current trail $t$, $LVM^{t}$ represents the up-to-date long valuable memory set up to trail $t$, $\theta$ represents the parameters of $SR_{lvm}$, mainly embodies through prompt.

Upon completion of the experience and failure lesson summarization processes, the reflection summary submodule $SR_{\text{sum}}$ performs \textbf{reflection summarization}, consolidating successful and failed experiences with key mentor guidance $M(q)$ to generate an improved plan.
The reflected new constructive plan will be stored in the $sm$ module and guide the planning process of $M_{actor}$ in the next trail. 
The comprehensive construction process can be described as:
\begin{equation}
Plan^{t}=SR_{sum}(G,\tau^{t},Exp^{t},Lesson^{t},M(G,\tau^{t}),\theta)
\label{constructivism} 
\end{equation}
where $G$ represents the task goal, $\tau^{t}$ represents the trajectory of current trail $t$, $Exp^{t}$ and $Lesson^{t}$ respectively represents the valuable experience and the precisely required failure lesson summarized through trail $t$, $M(q)$ represents the key knowledge from mentor, $\theta$ represents the parameters of $SR_{sum}$, mainly embodies through prompt.

\begin{figure*}[!htbp]
\vspace{-1mm}
    \centering
    \includegraphics[width=0.99\textwidth]{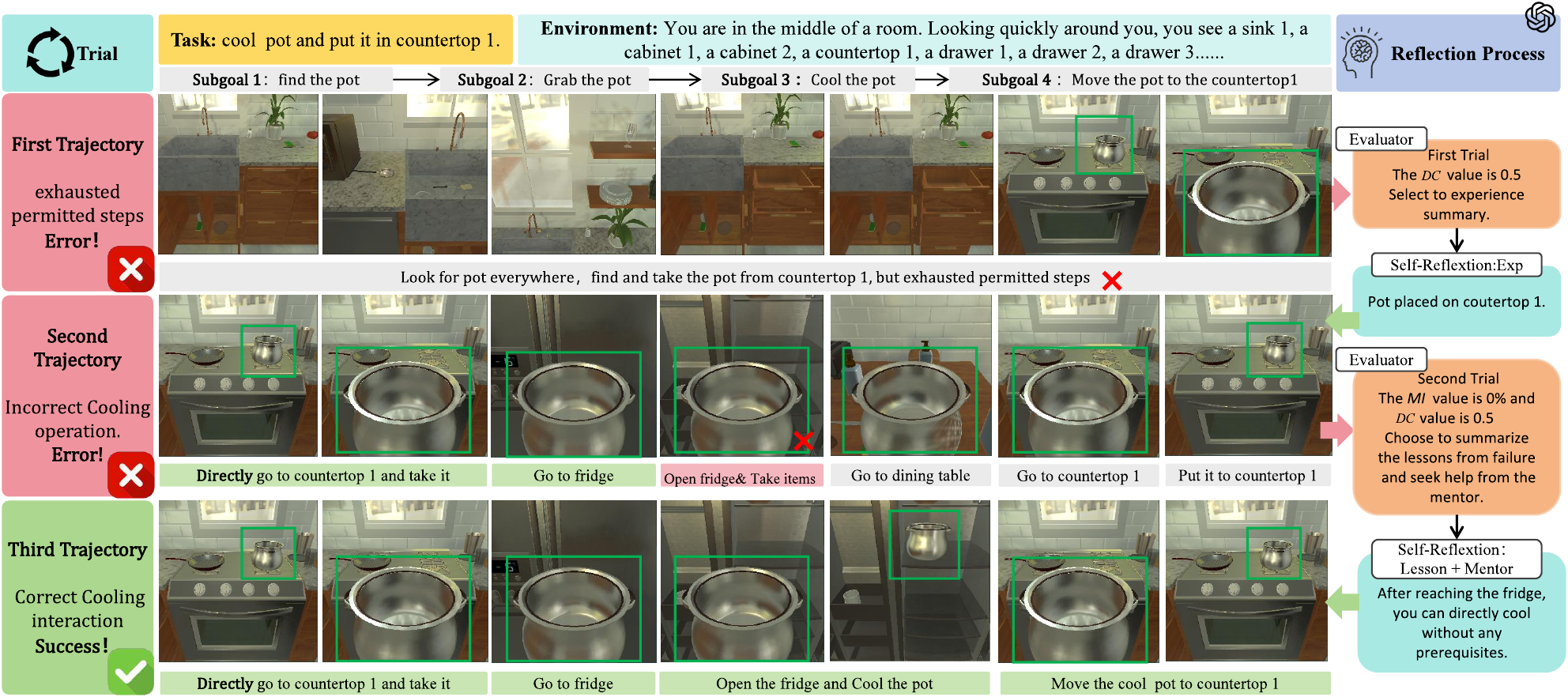}
    \caption{In an ALFWorld example, FISRF corrects a failed trajectory by summarizing experiences in the first episode, subsequently learning from failures and seeking mentor guidance in the second episode.}
    \label{fig_sim_example}
 \vspace{-5mm}
\end{figure*}

\subsection{The Flexible Reflection Process Under Mentor}
 A key innovation of our framework is the flexible, mentor-guided reflection process guided by difficulty of correction rather than difficulty of task itself. We assess this difficulty of correction based on task objectives, environmental context, and historical execution records, formulated as:
\begin{equation}
DC^t = \left(1 - \frac{Rate_{obj}^t + Rate_{inter}^t}{2}\right) 
\label{eq:difficulty_of_correction}
\end{equation}
Where $Rate_{obj}^t$ represents the task completion rate of the current trial $t$, defined as $\mathrm{Rate}_{\mathrm{obj}}^t = \frac{\mathrm{Obj}_\mathrm{traj}^t}{\mathrm{Obj}_\mathrm{all}}$, where $\mathrm{Obj}_\mathrm{all}$ represents the total number of sub-goals that the task needs to complete.$Rate_{inter}^t$ represents the current attempted item reach rate, defined as  $\mathrm{Rate}_{\mathrm{inter}}^t = \frac{\mathrm{Inter}_\mathrm{Traj}^t}{\mathrm{Inter}_\mathrm{all}}$, where $\mathrm{Inter}_\mathrm{all}$ represents the total number of items needed to be reached for the task.
Concurrently, we define the magnitude of improvement based on difficulty of correction, quantifying the improvement between consecutive attempts on the same task as:
\begin{equation}
MI^{t} = \frac{DC^{t-1} - DC^{t}}{1 - DC^{t-1}} \times 100\%
\label{eq:magnitude_of_improvement}
\end{equation}
where $DC^{t-1}$ and $DC^{t}$ denote the difficulty of correction in the current and previous rounds, respectively.
As illustrated in Figure~\ref{FISRF_fig3},the reflection strategy at trial $t$ is determined by $t$, $DC$ and $MI^{t}$: valuable experience summarization ($Exp^{t}$) triggers at Episode $=1$ or $MI^{t}>50\%$; failure lesson summarization ($Lesson^{t}$) triggers at Episode $>1$ and $MI^{t}\leq 50\%$; When Episode $=1$ with $DC>0.5$ or Episode $>1$ with $MI^{t}\leq 50\%$, the Agent is in the Zone of Proximal Development(ZPD), and mentor guidance $M(G,\tau^{t})$ is triggered.

\section{EXPERIMENTS}

In this section, we evaluate FISRF in the household simulation environment AlfWorld~\cite{shridhar2020alfworld} and further validate its effectiveness through real-world robotic experiments. Empirical results demonstrate that the proposed framework consistently outperforms baseline methods, achieving superior overall success rates as well as improvements across fine-grained performance metrics.

\subsection{Experimental Setup}



\textbf{Environment and Dataset.} We conduct our evaluation in \textbf{AlfWorld}, a text-based virtual household environment comprising five task categories: \textit{Put}, \textit{Clean}, \textit{Heat}, \textit{Cool}, and \textit{Examine}. Our experiments span the complete set of 134 tasks across these categories. Each task is evaluated over five planning epochs to ensure statistical robustness and enhance the generalizability of the results.

\textbf{Compared Methods.}
We benchmark FISRF against four representative categories of baselines in robotic planning and reflection:
1. \textbf{Plan-Only.} \textit{ProgPrompt}~\cite{singh2023progprompt}, which directly generates executable action sequences with LLMs under the in-context learning paradigm~\cite{dong2022survey}, without explicit reasoning or reflection mechanisms.
2. \textbf{Reason-Only.} \textit{ReAct}~\cite{yao2022react}, which interleaves reasoning and action generation, enabling the LLM to iteratively decide subsequent actions through step-wise deliberation.
3. \textbf{Reason-Reflect.} \textit{Reflexion}~\cite{shinn2024reflexion}, which augments ReAct with a reflection module that leverages environmental feedback to analyze failures and refine future planning.
4. \textbf{Reason-Flexible Reflect.} \textit{FCRF}~\cite{song2025fcrf}, which further incorporates valuable historical experiences and adaptively adjusts reflection intensity according to task complexity, enabling more flexible and targeted self-reflection.

\begin{table*}[t]
    \centering
    \small
    \setlength{\tabcolsep}{12pt}
    \vspace{3mm} 
    \caption{Success rates (\%) of FISRF and baseline methods on ALFWorld across different task categories. Results are reported under Gemini 2.5 and GPT-4o model}
    \label{Table:main1}
    \begin{tabular}{l|c c c c c |c c c}
        \toprule
        \multirow{2}{*}{\textbf{Method}} & \multicolumn{8}{c}{\textbf{Success Rate (\%)}$\uparrow$} \\
        \cmidrule(lr){2-9}
        & \textbf{Put} & \textbf{Clean} & \textbf{Heat} & \textbf{Cool} & \textbf{Examine} & \textbf{SR@1} & \textbf{SR@2} & \textbf{Overall} \\
        \midrule
        \multicolumn{9}{c}{\textit{Model: Gemini 2.5}} \\
        \midrule
        Plan-Only           & 82.9 & 71.0 & 82.6 & 90.5 & 66.7 & 73.9 & 76.1 & 79.1 \\
        Reason-Only         & 90.2 & 80.6 & 79.2 & 95.7 & 55.6 & 76.1 & 82.1 & 83.6 \\
        Reason-Reflect      & 97.6 & 80.6 & 78.3 & 90.5 & 66.7 & 76.1 & 81.3 & 85.1 \\
        Reason-Flex Reflect & \textbf{100.0} & 87.1 & 60.9 & 90.5 & \textbf{88.9} & 78.4 & 85.8 & 87.3 \\
        \rowcolor{gray!20}
        \textbf{FISRF (Ours)} & \textbf{100.0} & \textbf{90.3} & \textbf{91.3} & \textbf{100.0} & \textbf{88.9} & \textbf{91.0} & \textbf{93.3} & \textbf{94.8} \\
        \midrule
        \multicolumn{9}{c}{\textit{Model: GPT-4o}} \\
        \midrule
        Plan-Only           & 64.5 & 66.6 & 30.0 & 62.5 & 26.6 & 61.9 & 65.7 & 68.7 \\
        Reason-Only         & 86.3 & 87.5 & 77.1 & 93.3 & 58.3 & 80.6 & 81.3 & 82.8 \\
        Reason-Reflect      & 84.9 & 84.3 & 80.0 & 93.3 & 58.3 & 80.6 & 82.8 & 83.6 \\
        Reason-Flex Reflect & 91.4 & 90.3 & 93.5 & \textbf{100.0} & 63.6 & 82.1 & 89.0 & 91.0 \\
        \rowcolor{gray!20}
        \textbf{FISRF (Ours)} & \textbf{100.0} & \textbf{93.5} & \textbf{95.7} & \textbf{100.0} & \textbf{77.8} & \textbf{85.8} & \textbf{91.8} & \textbf{94.8} \\
        \bottomrule
    \end{tabular}
\end{table*}

\begin{table*}[t]
    \small 
    \centering
    \setlength{\tabcolsep}{1.5mm}
    \caption{Flexibility and Efficiency of reflection methods. Our FISRF demonstrates significantly higher flexibility and efficiency. }
\tabcolsep=0.012\linewidth
\begin{tabular}{l|cccc|cc}
    \toprule
    \multirow{2}{*}{\textbf{Methods}} & 
    \multicolumn{4}{c}{\textbf{Efficiency}} &
    \multicolumn{2}{c}{\textbf{Flexibility}} \\
    \cmidrule(lr){2-5} \cmidrule(lr){6-7}
    & $\mathrm{Cost}_{\mathrm{ref}}$ @4 $\downarrow$ & $\mathrm{Rate}_{\mathrm{rev}}(\%)$ @4  $\uparrow$ &  $\mathrm{Round}_{\mathrm{rev}}$ @2  $\downarrow$ & $\mathrm{Round}_{\mathrm{rev}}$ @4 $\downarrow$ 
    & {$\text{R}$ $\uparrow$} & {$\text{CV}(\%)$ $\uparrow$} \\
    \midrule
    \multicolumn{7}{c}{${Model: Gemini}$ ${2.5}$ } \\ 
    \midrule
    Reason-Reflect  &  913.7 & 64.3  & 3.26 & 4.79 & 207 & 23.5 \\
    Reason-Flex Reflect & 558.3 & 69.6 & 2.30 & 3.15 & 237 & 21.5 \\
    \rowcolor{gray!20} 
    \textbf{FISRF(Ours)}  & \textbf{370.5} &  \textbf{87.5} & \textbf{1.45} & \textbf{1.71} & \textbf{253} & \textbf{36.7}\\
    \midrule
    \multicolumn{7}{c}{${Model: GPT}$ ${4o}$ } \\ 
    \midrule
    Reason-Reflect  & 953.6 & 47.6 & 3.58 & 5.70 & 175 & 22.4 \\
    Reason-Flex Reflect & 570.4 & 71.4 & 2.44 & 3.13 & 197 & 18.2 \\
    \rowcolor{gray!20} 
    \textbf{FISRF(Ours)}  & \textbf{427.3} &  \textbf{87.5} & \textbf{1.79} & \textbf{2.14} & \textbf{247} & \textbf{24.3}\\
    \bottomrule
\end{tabular}
    \label{Table:main2}
\end{table*}

\textbf{Metrics.} We evaluate the methods from three aspects:
\begin{itemize}
    \item \textbf{Success Rate} quantifies the overall effectiveness of the planning and reflection mechanisms, defined as the proportion of tasks successfully completed by the end of each trial.    
    \item \textbf{Efficiency} is assessed through two indicators: \textbf{Reflection Cost} ($\mathrm{Cost}_{\mathrm{ref}}$): the average token cost per successful recovery, defined as $\mathrm{Cost}_{\mathrm{ref}} = \frac{\mathrm{Cost}_{\mathrm{all}}}{\mathrm{E}_{\mathrm{corrected}}}$,where $\mathrm{Cost}_{\mathrm{all}}$ is the total token consumption across all episodes and $\mathrm{E}_{\mathrm{corrected}}$ is the number of tasks recovered via reflection. \textbf{Revision Rate} ($\mathrm{Rate}_{\mathrm{rev}}$): the fraction of initially failed tasks that are successfully recovered, $\mathrm{Rate}_{\mathrm{rev}} = \frac{\mathrm{E}_{\mathrm{corrected}}}{\mathrm{E}_{\mathrm{initial}}}$, where $\mathrm{E}_{\mathrm{initial}}$ denotes the number of failures in the initial planning phase. \textbf{Revision Round} ($\mathrm{Round}_{\mathrm{rev}}$): the mean number of reflection iterations required to successfully recover from a failed task, defined as ($\mathrm{Round}_{\mathrm{rev}}$) = $\frac{\mathrm{Round}_{\mathrm{all}}}{\mathrm{E}_{\mathrm{corrected}}}$, where $\mathrm{Round}_{\mathrm{all}}$ the total number of reflection rounds accumulated across all episodes and $\mathrm{E}_{\mathrm{corrected}}$ is the number of tasks recovered via reflection. 
    \item \textbf{Flexibility} evaluates the adaptability of reflection length to task complexity, quantified by: \textbf{Range} (R): The difference between the maximum number of tokens in a single reflection and the minimum number of tokens in a single reflection. \textbf{Coefficient of Variation} (CV): normalized dispersion, defined as $\text{CV} = \frac{\mathrm{Cost}_{\mathrm{ref}, \text{STD}}}{\mathrm{Cost}_{\mathrm{ref}, \text{AVE}}}$ 
\end{itemize}

\subsection{Main Results}

The main results are reported in Table~\ref{Table:main1} and Table~\ref{Table:main2}. To balance performance and computational cost, all experiments employ the official \textit{GPT-4o} and \textit{Gemini 2.5} models. Figure~\ref{fig_sim_example} presents a representative long-horizon task solved by FISRF.
The key findings are summarized as follows:
1. FISRF achieves a 100\% success rate on both the \textit{Put} and \textit{Cool} tasks. Compared with baseline methods, it yields performance gains of over 30\% on the \textit{Examine} task and more than 4\% overall (Table~\ref{Table:main1}), demonstrating substantially enhanced reflection capability. These results indicate that combining failure analysis, correction-difficulty estimation, and mentor guidance improves recovery across heterogeneous task types. 2. We observe no significant correlation between reflection token length and subsequent error-correction effectiveness. Moreover, reduced flexibility and efficiency in self-reflection consistently correspond to lower success rates. These observations support the necessity of adaptive reflection and validate the design of our proposed evaluation metrics. 3. Among all methods, the \textbf{Plan-Only} baseline yields the lowest success rate, highlighting the limitations of direct action generation without reasoning or reflection. \textbf{Reason-Only} and \textbf{Reason-Reflect} achieve stronger performance; however, although \textbf{Reason-Reflect} generally surpasses \textbf{Reason-Only}, it underperforms on certain tasks. This suggests that fixed-template reflection does not universally translate into performance gains. 4. In contrast to prior approaches, FISRF dynamically adjusts reflection intensity over a broader and more variable range (Table~\ref{Table:main2}). The metric $R$ increases by over 6.8\%, and the coefficient of variation (CV) rises by more than 8.5\%, demonstrating enhanced \textbf{flexibility}. In terms of \textbf{efficiency}, FISRF achieves a higher revision rate and lower reflection cost, reflecting stronger error-correction capability and improved cost-effectiveness. Specifically, $\mathrm{Rate}_{\mathrm{ref}}$ improves by more than 22.5\% and $\mathrm{Cost}_{\mathrm{rev}}$ decreases by over 25.1\%, while $\mathrm{Round}_{\mathrm{rev}}$ decreases by over 31.6\%. Overall, FISRF adaptively allocates computational resources, effectively leverages diverse execution feedback, and achieves superior performance–cost trade-offs.

\begin{figure*}[!htbp]
      \centering
      \includegraphics[width=1.0\textwidth]{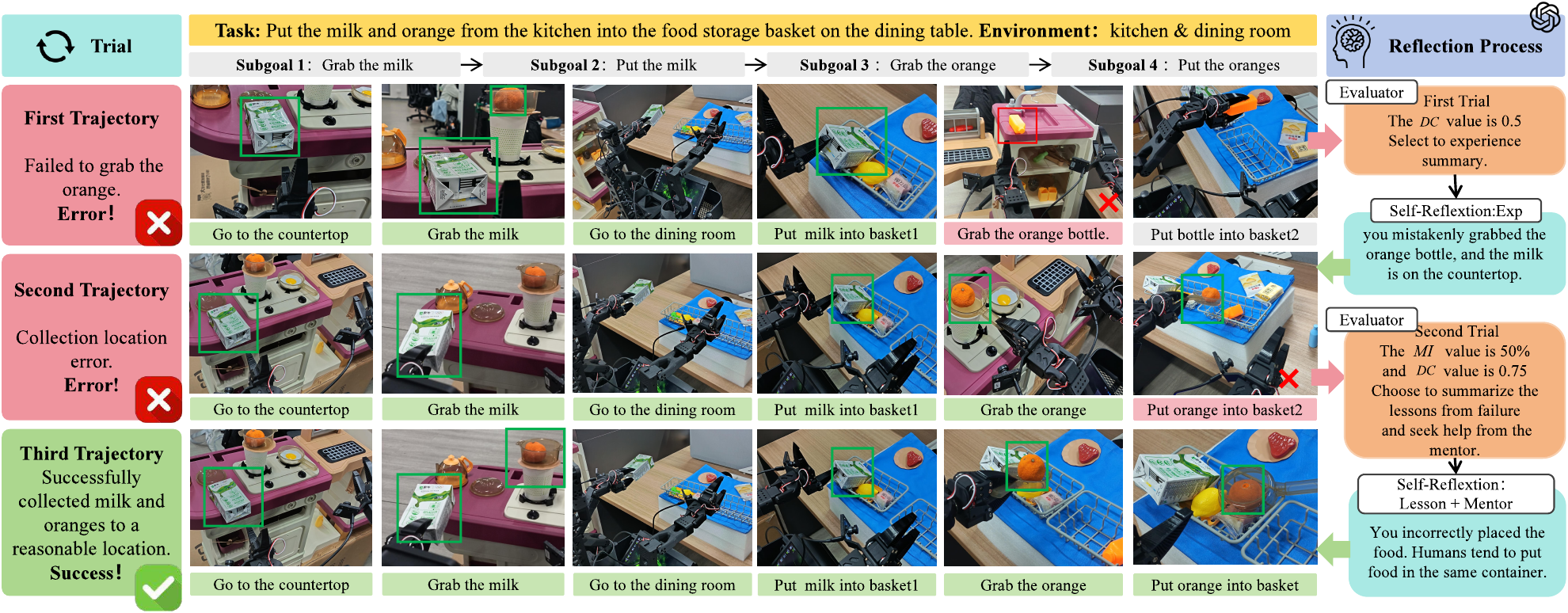}
      \caption{Food collection task of a failed task successfully refined using our FISRF.}
    \label{real_world}
 \vspace{-6mm}
\end{figure*} 
\section{ANALYSIS AND DISCUSSION}

\subsection{Episode Analysis of Self-Reflection Process}

\begin{figure}
      \centering
      
      \includegraphics[width=0.5\textwidth]
    {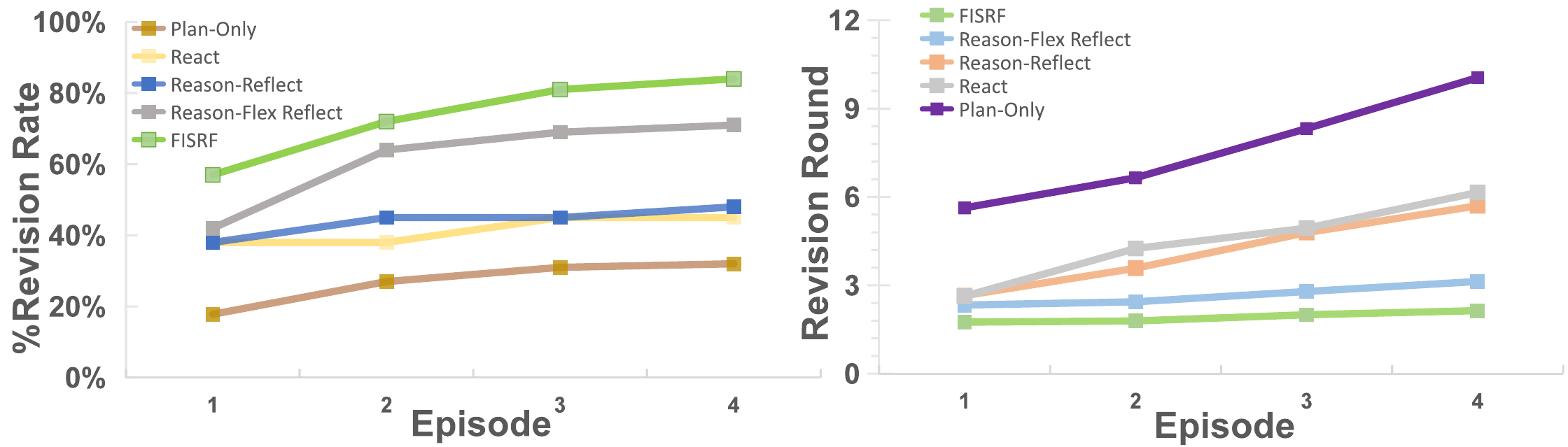}
      \caption{Result of success rate and round revision with different episodes. Our FISRF outperforms in all episodes, converges more quickly and has a significant better efficiency. }
    \label{EpisodeAnalysis}
    \vspace{-5mm}
\end{figure}

We conduct an episode-level analysis of iterative self-reflection, as shown in Figure~\ref{EpisodeAnalysis}. For each initially failed long-horizon task, the LLM replans over five episodes, with task-completion outcomes recorded after each episode. The main observations are as follows.

1. FISRF consistently outperforms all baselines and converges faster, demonstrating stronger robustness and error-recovery capability with fewer replanning iterations.

2. \textbf{Plan-Only} exhibits the weakest recovery performance. Although \textbf{Reason-Only} and \textbf{Reason-Reflect} improve task success, their marginal difference suggests that fixed-pattern reflection provides limited additional benefits.

3. As shown in the right panel, mentor-provided knowledge enables FISRF to resolve errors with fewer reflection iterations, improving reflection efficiency and convergence speed.

Overall, these results demonstrate that adaptive reflection and the integration of successful experiences, failure-derived lessons, and mentor guidance facilitate more efficient error correction and iterative refinement.








\subsection{Ablation Study}

To comprehensively evaluate the contribution of each module and examine their interactions, we conduct ablation studies on the full ALFWorld dataset. The detailed results are reported in Table~\ref{Table:ablation}.
In the \textbf{w/o Exp} variant, removing experience retention reduces the overall success rate by 7.9\% and decreases $R$ by 21.7\%. This confirms that accumulated experience is critical for reflective adaptability and stable long-horizon planning. In the \textbf{w/o Fail} model, the variant that excludes lessons derived from failures results in a 6.3\% decline in success rate, a 39.7\% reduction in CV, and a 27.5\% increase in reflection cost. These results underscore the importance of failure-informed knowledge in preserving reflective flexibility and ensuring cost-effective recovery in long-horizon planning.

\begin{table}[htbp]
    \centering
    \small
    \setlength{\tabcolsep}{2pt}
    \vspace{2.5mm} 
    \caption{Ablation study of FISRF components in AlfWorld tasks. All designed modules contribute to the performance.}
    \begin{tabular}{@{}lcccccccc@{}}
        \toprule
        \textbf{Method} & \textbf{SR } & \textbf{R} & \textbf{CV} & $\mathbf{Round_{rev}}$ & $\mathbf{Cost_{ref}}$ & $\mathbf{Rate_{rev}}$ \\
        & (\%) & ($\uparrow$) & ($\uparrow$) & ($\downarrow$) & ($\downarrow$) & ($\uparrow$) \\
        \midrule
        \textbf{Full}           & \textbf{94.8} & \textbf{253} & \textbf{36.7} & \textbf{1.7} & \textbf{370.5} & \textbf{87.5} \\
        w/o Exp               & 87.3 & 198 & 24.5  &  3.6 & 487.3 & 68.7 \\
        w/o Lesson           &  88.8 & 208 & 22.1 & 2.4 & 472.5 & 75.3 \\
        w/o Mentor              & 82.9 & 272 & 27.6 & 4.2 & 550.7 & 63.5 \\
        \bottomrule
    \end{tabular}
    \label{Table:ablation}
 \vspace{-5mm}
\end{table}

In the \textbf{w/o Mentor} model, removing mentor guidance reduces the success rate by 12.6\% and increases reflection cost by 48.6\%, demonstrating its importance in improving performance and efficiency. Ablating any individual module further degrades flexibility and efficiency, leading to a lower overall success rate. 


\subsection{Real-World Robotic Experiment}
We validate the proposed approach on a dual-arm mobile robot platform in real-world kitchen item manipulation tasks. As illustrated in Figure~\ref{real_world}, the robot equipped with FISRF demonstrates enhanced self-reflection and recovery performance. Additional details will be provided on our website and in supplementary videos.


\section{CONCLUSIONS}
We presented \textbf{FISRF}, a mentor-guided reflection framework that operationalizes the robot's \textbf{Zone of Proximal Development(ZPD)} through difficulty of correction and magnitude of improvement. FISRF adaptively combines autonomous experience, failure lessons, and retrieved mentor knowledge to improve subsequent planning. Experiments in ALFWorld and on real robotic platforms demonstrate consistent gains in task success, correction efficiency, and reflection flexibility. Future work will investigate automatic knowledge-base expansion and the transfer of mentor guidance across environments and robot platforms.

\section{Acknowledgement}
This research was supported by The Science and Technology Program of Zhejiang Province (Grant No. 2025C02038) and Key R\&D Program of Zhejiang (Grant No. 2026SDXT007).






\bibliographystyle{unsrt}
\bibliography{reference}

\end{document}